%% file: mmfashion.tex
\documentclass[journal]{IEEEtran}
\usepackage{cite}

\ifCLASSINFOpdf
  \usepackage[pdftex]{graphicx}
  \graphicspath{{../pdf/}{../jpeg/}}
  \DeclareGraphicsExtensions{.pdf,.jpeg,.png}
\else
  \usepackage[dvips]{graphicx}
  \graphicspath{{../eps/}}
  \DeclareGraphicsExtensions{.eps}
\fi

\usepackage{amsmath}
\usepackage{amssymb}
\usepackage{algorithm}
\usepackage{algorithmic}
\usepackage[T1]{fontenc} 
\usepackage{multirow}
\usepackage{siunitx}
\usepackage{color}
\usepackage{parnotes}
\usepackage{booktabs}
\usepackage{caption}
\usepackage{xcolor}
\usepackage{xspace}
\usepackage{tikz}

\usepackage[pagebackref=true,breaklinks=true,letterpaper=true,colorlinks,bookmarks=false]{hyperref}

\begin{document}
%
\title{MMFashion: An Open-Source Toolbox for \\ Visual Fashion Analysis}
%
%
%

\author{Xin~Liu,
        Jiancheng~Li,
        Jiaqi~Wang,
        and~Ziwei~Liu 
        \\ ~ \\
        Multimedia Lab, The Chinese University of Hong Kong 
}


\twocolumn[{
\renewcommand\twocolumn[1][]{#1}%
\maketitle
\vspace{-30pt}
\begin{center}
  \centering
  \includegraphics[width=0.9\textwidth]{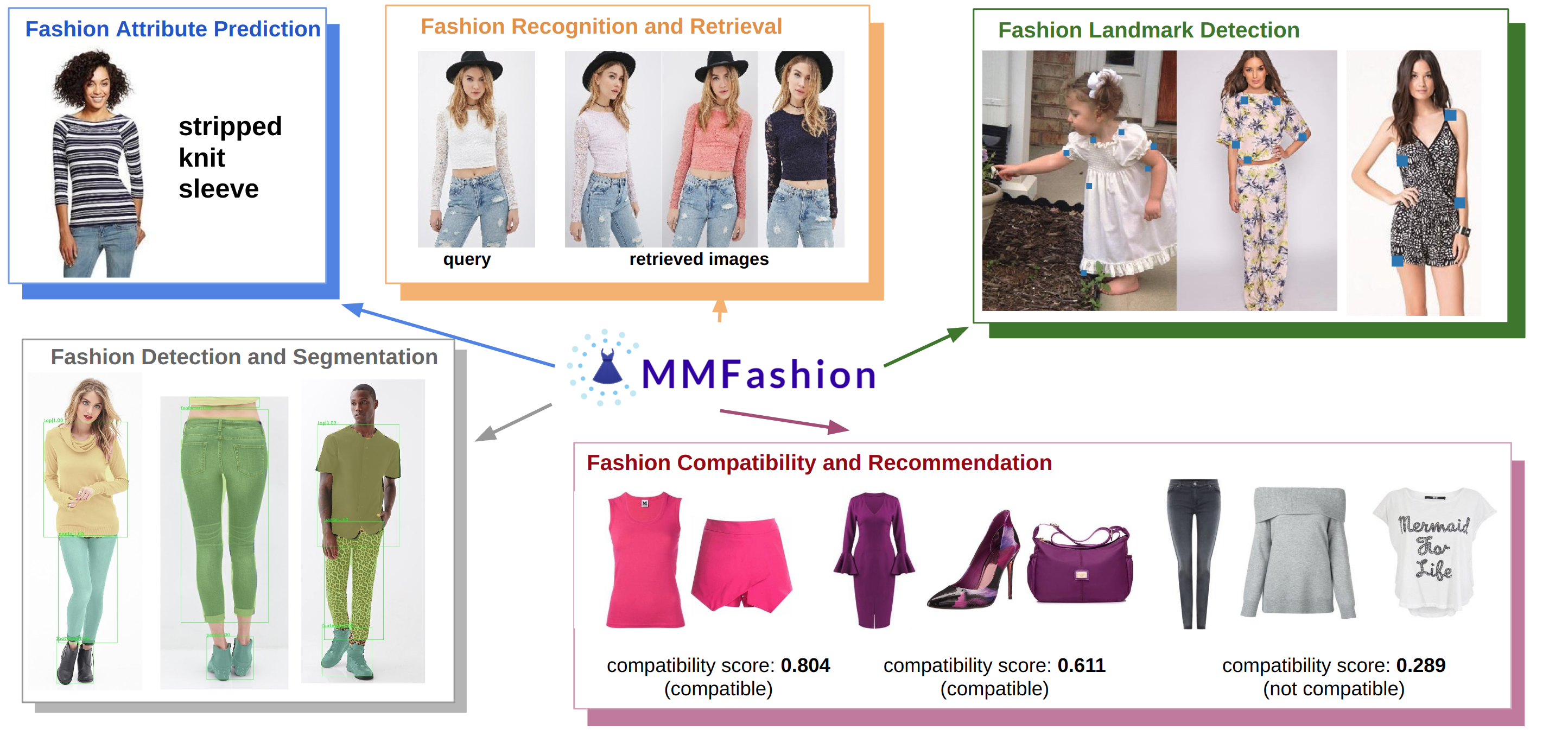}
  \vspace{-2pt}
 \captionof{figure}{MMFashion, a comprehensive, flexible and user-friendly open-source toolbox. It is currently the most complete platform for visual fashion analysis in deep learning community, whose users include not only computer vision researchers, but also layman users.}
  \label{fig:general}
\end{center}
}]


\IEEEpeerreviewmaketitle

\input{0_abstract.tex}
\input{1_introduction.tex}
\input{2_main_features.tex}
\input{3_applications.tex}

\ifCLASSOPTIONcaptionsoff
  \newpage
\fi


\bibliographystyle{IEEEtran}


\end{document}

%% file: 0_abstract.tex
\begin{abstract}
We present MMFashion, a comprehensive, flexible and user-friendly open-source visual fashion analysis toolbox based on PyTorch. 
This toolbox supports a wide spectrum of fashion analysis tasks, including \textit{Fashion Attribute Prediction}, \textit{Fashion Recognition and Retrieval}, \textit{Fashion Landmark Detection}, \textit{Fashion Parsing and Segmentation} and \textit{Fashion Compatibility and Recommendation}. It covers almost all the mainstream tasks in fashion analysis community. 
MMFashion has several appealing properties.
Firstly, MMFashion follows the principle of modular design. The framework is decomposed into different components so that it is easily extensible for diverse customized modules.   
In addition, detailed documentations, demo scripts and off-the-shelf models are available, which ease the burden of layman users to leverage the recent advances in deep learning-based fashion analysis. 
Our proposed MMFashion is currently the most complete platform for visual fashion analysis in deep learning era, with more functionalities to be added. This toolbox and the benchmark could serve the flourishing research community by providing a flexible toolkit to deploy existing models and develop new ideas and approaches. We welcome all contributions to this still-growing efforts towards open science: \url{https://github.com/open-mmlab/mmfashion}.
\end{abstract}

%% file: 1_introduction.tex
\begin{table*}[h!]
    \scriptsize
    \centering
    \caption{Comparison of different high-starred fashion analysis toolkits: plenty of open-source codebases target on particular aspects of fashion analysis tasks, while none of them comprehensively support multiple tasks.}
    \begin{tabular}{c c c c c c}
    \toprule
    {} & {Attribute Prediction} & {Cloth Retrieval} & {Landmark Detection} & {Detection and Segmentation} & {Compatibility and Recommendation} \\
    \midrule
    Fashion-MNIST\cite{xiao2017/online}  & \checkmark &  &  &  &  \\
    tf.fashionAI\cite{tfFashionAI}   &       &  & \checkmark & & \\
    fashion-detection\cite{liu2016deepfashion}  &  & & & \checkmark & \\
    deep-fashion-retrieval\cite{deepFashionRetrive} & & \checkmark & & & \\
    fashion-landmarks\cite{liu2016fashion}  & & & \checkmark &  & \\
    fashion-recommendationv\cite{fashionRecommend}  & & \checkmark & & & \checkmark \\
    polyvore\cite{han2017learning} & & & & & \checkmark \\
    MMFashion& \checkmark & \checkmark & \checkmark & \checkmark & \checkmark \\
    \bottomrule
    \end{tabular}
    \label{tab:comparison}
\end{table*}

\section{Introduction}
Fashion plays an important role in human society. People actively engage in fashion-related activities as means of self-expressing and social connection~\cite{liu2020learning, shi2019learning}. 
With the rapid growth of e-commerce, not only the professional researchers, but also the layman users tend to seek assistance from intelligent visual fashion analysis tools. 
Such enthusiasm further motivates the evolution of combining fashion with deep learning toolkit. Numerous fashion-related tasks, e.g., clothes attribute prediction, fashion recognition and retrieval, fashion parsing and segmentation, fashion compatibility analysis and recommendation, have been widely applied in online shopping platforms. 
These state-of-the-art algorithms are being pushed forward by worldwide researchers in a day-by-day pace. 
Though intelligent fashion industry is highly developed, there exists no unified platform or codebase to integrate all these diverse fashion analysis tools compared with other subfields such as object detection and video action analysis, shown as Table~\ref{tab:comparison}. 
Such dilemma not only hinders the steps of professional researchers in advancing algorithms, but also discourages layman users in involving artificial intelligence to their daily life.

In order to bridge the aforementioned gaps, we propose \textbf{MMFashion}, a comprehensive, flexible and user-friendly open-source visual fashion analysis tool based on PyTorch \cite{paszke2017automatic}.
MMFashion has several appealing main features:
\textbf{1) Supporting Diverse Tasks.}
We support a wide spectrum of fashion analysis tasks, which covers almost all the mainstream tasks in current research community. Together with each task, we also include popular datasets and representative methods. 
\textbf{2) Modular Design.}
We decompose the entire framework into several re-configurable modules. With this manner, developers can easily add or reconstruct a customized fashion analysis pipeline by combining different modules. Our provided configuration files are easy to understand even for layman users.
\textbf{3) Detailed Benchmark Study.} 
We construct a comprehensive model zoo and measure the accuracies of individual models on a common ground. In this way, users can effortlessly download the corresponding pretrained models according to their application and accuracy requirement. At the meanwhile, professional researchers can refer to our benchmark study to evaluate their model performance.

In the following of this paper, we will first introduce the supported tasks and application scenarios, then highlight the main features of our codebase, and finally present the benchmark studies. We hope our proposed MMFashion can 1) serve as a unified platform to push forward the research efforts in fashion community, and also 2) benefit layman users to leverage the recent advances in visual fashion analysis.

\section{Supported Tasks}
\subsection{Fashion Attribute Prediction}
Fashion Attribute Prediction predicts the attributes of the cloth, e.g., print, T-shirt, blouse, etc. 
Such attribute prediction belongs to multi-label prediction problems. Following \cite{gong2013deep}, the top-k recall rate is adopted as the measuring criteria.

\subsection{Fashion Recognition and Retrieval}
Fashion Recognition and Retrieval aims at determining whether two images belong to the same clothing item \cite{liu2016deepfashion}. The two images of an image-pair can be taken from different sources. ``In-shop Clothes Retrieval'' indicates both images are in-shop. ``Consumer-to-Shop Clothes Retrieval`` means that one images is provided by the consumer and the other one is an in-shop counterpart. This task is of great use in e-commence scenarios nowadays.

\subsection{Fashion Landmark Detection}
Fashion landmarks are key-points located at the functional region of clothes, e.g., neckline, hemline and cuff \cite{liu2016deepfashion}.
The positions of landmarks defined on the fashion items are detected in this task.

\subsection{Fashion Parsing and Segmentation}
Object detection and instance segmentation are fundamental techniques in computer vision area. Fashion Parsing and Segmentation is specially designed for detection and segmentation on cloth data.

\subsection{Fashion Compatibility and Recommendation}
This task is composed of two main functions: 1) determining the compatibility relationship of a given fashion outfit; 2) recommending an item that matches with the existing set \cite{han2017learning}.

%% file: 2_main_features.tex
\section{Architecture}
\subsection{Model Representation}

Although the model architectures of different tasks vary from each other, they have common components, which can be roughly summarized into the following classes.

\textbf{Backbone.} Backbone is the part that transforms an image to feature maps, such as a ResNet-50 \cite{he2016deep} without the last fully connected layer.

\textbf{Head.} Head is the part that takes the features as input and makes task-specific predictions, such as attribute predictor, clothes retriever and landmark regressor.

\begin{figure}
\vspace{-10pt}
\begin{center}
  \includegraphics[width=0.8\linewidth]{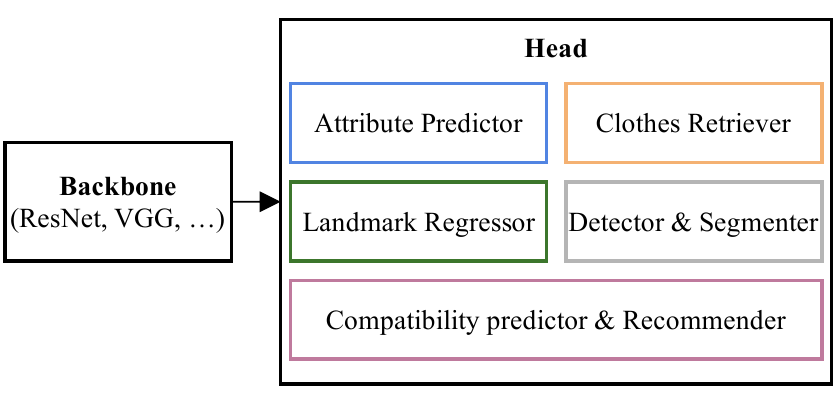}
\end{center}
\vspace{-10pt}
  \caption{Framework of MMFashion.}
\label{fig:framework}
\vspace{-10pt}
\end{figure}

With the above abstractions, the framework is illustrated in Fig.~\ref{fig:framework}. We can develop our own methods by simply creating some new components and assembling existing ones.

\subsection{Training Pipeline}

We use the unified training pipeline with hooking mechanism as shown in MMDetection \cite{chen2019mmdetection}. The training processes of these fashion analysis tasks share a similar workflow, where training epochs and validation epochs run iteratively and validation epochs are optional. In each epoch, we forward and backward the model by many iterations. To make the pipeline more flexible and easy to customize, we define a minimum pipeline which just forwards the model repeatedly. 
In order to run the training process of different fashion analysis tasks, we perform some self-defined operations before or after some specific steps, such as updating learning rate, saving checkpoint and evaluate the pretrained model. 

%% file: 3_applications.tex
\section{Benchmark Study}
\subsection{Dataset}
\noindent
\textbf{DeepFashion.}
DeepFashion is a large-scale clothes database, which contains over 800,000 diverse fashion images ranging from well-posed shop images to unconstrained consumer photos \cite{liu2016deepfashion}.
We use its corresponding benchmarks for attribute prediction, clothes retrieval, landmark detection respectively.
Additionally, we build the clothes detection and segmentation annotations as COCO-style based on ``In-Shop Clothes'' benchmark to serve \textit{Fashion Parsing and Segmentation} task. 
From our knowledge, this human-labeled segmentation benchmark, even strict with hairs, is the most accurate open-source dataset in fashion community at present. 

\noindent
\textbf{Polyvore.}
Polyvore contains rich multimodel information, e.g., images and text descriptions of fashion items, associated tags, popular score and type information, that makes it applicable for \textit{fashion compatibility learning}.
Here we use \cite{vasileva2018learning} reorganized version, that includes two split versions: an eaiser("Polyvore") one that allows overlapped items in train and test sets, and a difficult("Polyvore-D") one that ensures no garment appears in more than one set.


\begin{table}
    \caption{Mean Per-Attribute Prediction performance.}
    \scriptsize
    \begin{center}
    \begin{tabular}{c c c c c}
    \toprule
    {Backbone}  &  {Pooling} & {Loss} & {Top-5} & {Top-5} \\
    {}          &   {}       & {}     & {Recall(\%)}       & {Accuracy(\%)} \\
    \midrule     
    VGG16 &  Global Pooling & Cross-Entropy & 13.70 & \textbf{99.81} \\
    VGG16 &  Landmark Pooling & Cross-Entropy & 14.79 & 99.27 \\
    ResNet50 & Global Pooling & Cross-Entropy & 23.52 & 99.29 \\
    ResNet50 & Landmark Pooling & Cross-Entropy & \textbf{30.84} & 99.30 \\
    \bottomrule
    \end{tabular}
    \end{center}
    \label{tab:attribute}
    \vspace{-10pt}
\end{table}

\subsection{Evaluation Metrics}
\noindent
\textbf{Fashion Attribute Prediction.}
We use the standard top-k recall rate and accuracy \cite{gong2013deep} to evaluate the prediction model. For each cloth image, we assign $k$ highest-ranked attributes and compare the assigned attributes with the ground-truth attributes. We compute the recall rate and the accuracy for each attribute separately, shown as Equation~\ref{eq:attribute}:

\begin{equation}
\begin{split}
    per-attribute-recall =\frac{1}{c} \sum_{i=1}^{c} \frac{N_i^{tp}}{N_i^{g}} \\
    per-attribute-accuracy =\frac{1}{c} \sum_{i=1}^{c} \frac{N_i^{tp} + N_i^{tn}}{N_i^{p}}
\end{split}
\label{eq:attribute}
\end{equation}

where $c$ is the number of attributes, $N_i^{tp}$ is the number of true positive predicted image, $N_i^{tn}$ is the number of true negative predicted image for attribute $i$,  $N_i^{g}$ is the number of ground-truth attributes for attribute $i$, and $N_i^{p}$ is the number of predictions for attribute $i$. 



\begin{figure}
\vspace{-20pt}
\begin{center}
  \includegraphics[width=0.8\linewidth]{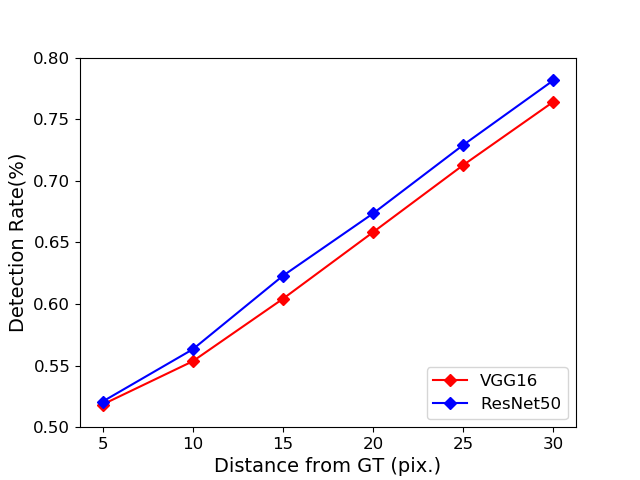}
\end{center}
\vspace{-5pt}
  \caption{Performance of fashion landmark detection on different distances from the ground truth. (px) represents pixels.}
\label{fig:landmark}
\vspace{-5pt}
\end{figure}

\begin{table}
    \caption{Fashion detection and segmentation results on different scales of images.}
    \begin{center}
    \begin{tabular}{c c c c}
    \toprule
    {IoU}  &  {area} & {$AP_{bbox}$} & {$AP_{mask}$} \\
    \midrule     
    0.50 &  all  & 0.815 & 0.759 \\
    0.75 &  all  & 0.663 & 0.637 \\
    0.50 : 0.95 & all & \textbf{0.599} & \textbf{0.584} \\
    \midrule
    0.50 : 0.95 & small & 0.118 & 0.071 \\
    0.50 : 0.95 & medium & 0.398 & 0.352 \\
    0.50 : 0.95 & large & 0.608 & 0.650 \\
    \bottomrule
    \end{tabular}
    \end{center}
    \label{tab:detect}
    \vspace{-10pt}
\end{table}

\begin{figure*}
\centering
\includegraphics[width=1.0\linewidth]{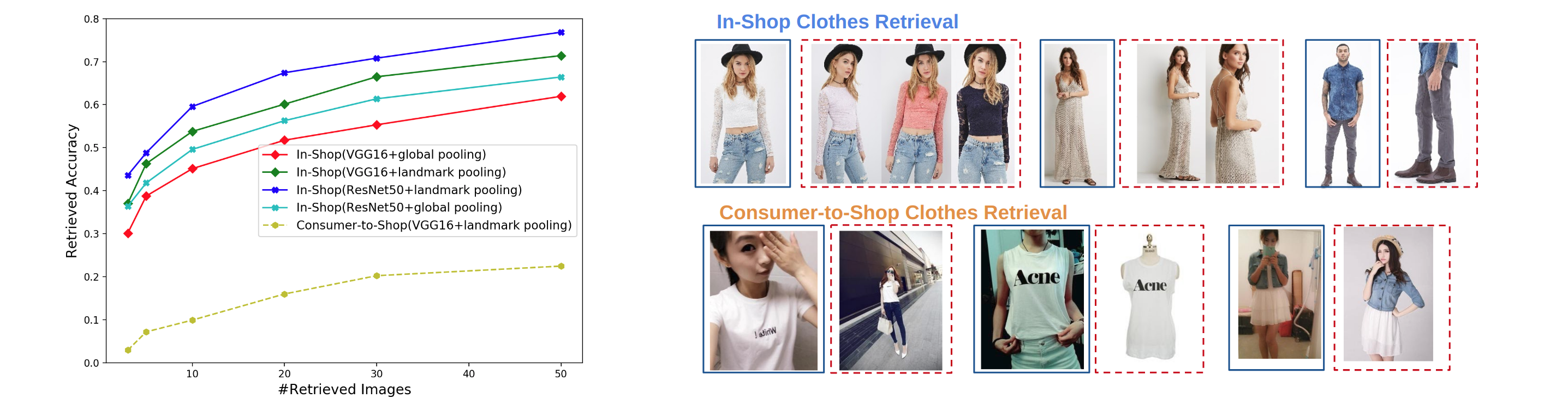}
\caption{Fashion Recognition and Retrieval. \emph{Left}: Retrieval accuracies of different methods on In-shop and Consumer-to-Shop Clothes Retrieval Benchmarks. \emph{Right}: Visual examples of cloth retrieval. Figures in blue squares are query images, in red squares are the retrieved images.}
\label{fig:retrieval}
\end{figure*}

\begin{table*}
\scriptsize
\caption{\emph{Left}: Fill-in-the-Blank(FITB) Accuracy and Fashion Compatibility Prediction AUC in two split versions of Polyvore dataset. \emph{Right}: Visual Results of these two tasks.}
\vspace{-5pt}
\parbox{0.4\linewidth}{
\centering
\begin{tabular}{c|c c c}
    \toprule
    {Dataset} & {Embedding} & {FITB} & {Compat.} \\
    {}  & {Projection} & {Acc.(\%)} & {AUC} \\
    \midrule
     Polyvore-D & fully-connected layer & 50.4  & 0.80\\
     Polyvore-D & learned metric & \textbf{55.6} & 0.84 \\
     Polyvore & fully-connected layer & 53.5 & \textbf{0.85} \\
     \bottomrule
\end{tabular}
}
\hfill
\parbox{0.6\linewidth}{
\centering
\includegraphics[width=1.0\linewidth]{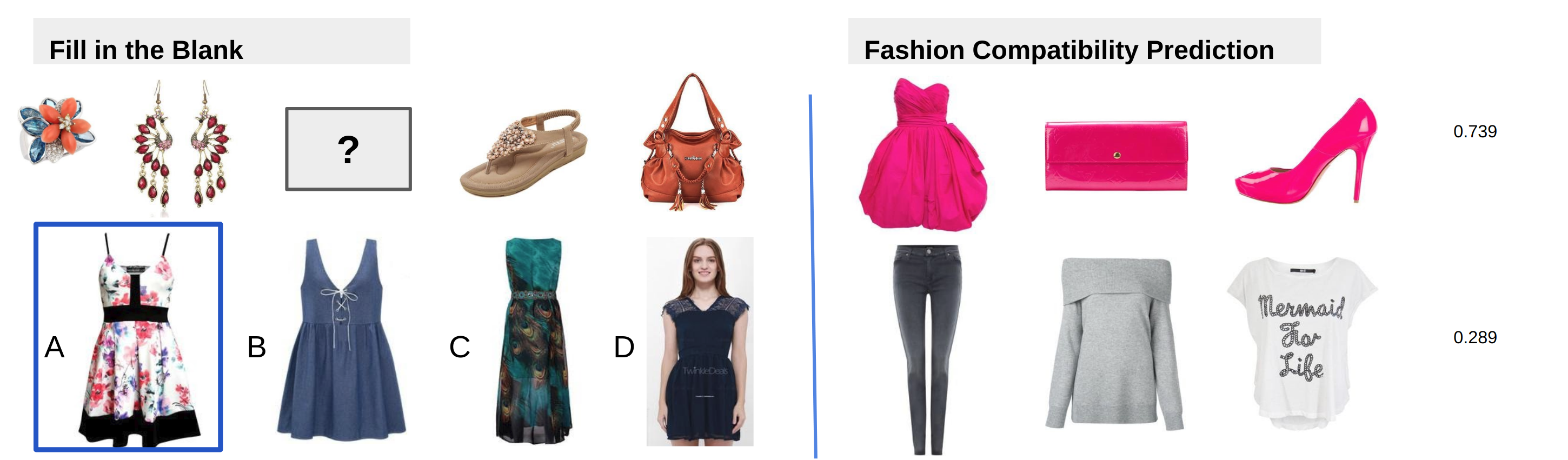}
}
\vspace{-4pt}
\label{table:learning_comp}
\end{table*}

\noindent
\textbf{Clothes Retrieval.}
Similarly, we deploy the top-k recall rate to evaluate the clothes retrieval model, shown as Equation~\ref{eq:attribute}. 

\noindent
\textbf{Landmark Detection.}
Normalized error (NE) \cite{liu2016fashion}, defined as $L2$ distance between the predicted landmarks and ground truth landmarks in the normalized coordinate space, is used to evaluate the fashion landmark detection, explained as Equation~\ref{eq:NE}. $dx$, $dy$ mean the distance between the predicted and the ground truth in $x$ and $y$ coordinates, $w$, $h$ are the image width and height. 
Considering some landmarks are invisible due to challenging body pose, we only count the visible landmarks.

\begin{equation}
    NE = \sqrt{ (\frac{dx}{w})^2 + (\frac{dy}{h})^2  }
    \label{eq:NE}
\end{equation}

\noindent
\textbf{Cloth Detection and Segmentation.}
We extend this function module using mmdetection codebase \cite{chen2019mmdetection}, that adopts standard evaluation metrics for COCO dataset and multiple IoU thresholds from 0.5 to 0.95 are applied. The detection and segmentation performance are measured with mean Average Precision(mAP).

\noindent
\textbf{Fashion Compatibility and Recommendation.}
Fill-in-the-Blank and Fashion-Compatibility-Prediction are utilized to evaluate the model performance, shown as Fig.~\ref{table:learning_comp}. 
\textit{Fill-in-the-Blank} gives a sequence of fashion items, and requests to select one item from four choices that is most compatible with those given items.
The performance is measured as the accuracy of correctly answered questions.
\textit{Fashion Compatibility Prediction} requests to score a candidate outfit in order to determine if they are compatible or not.
Performance is evaluated using the area under a receiver operating curve(AUC) \cite{han2017learning}.


\subsection{Benchmarking Results}

\noindent
\textbf{Fashion Attribute Prediction.}
We reimplement the method in~\cite{liu2016deepfashion} to predict 1000 attributes per image and report the top5 mean per-attribute recall rate and accuracy in Table~\ref{tab:attribute}.
The ``landmark pooling'' method is to extract and to utilize local feature maps around the landmark $l$, while the basic ``global pooling'' simply uses the global features of the whole image.

\noindent
\textbf{Fashion Recognition and Retrieval.}
We complete two subtasks here, in-shop cloth retrieval and consumer-to-shop cloth retrieval, shown in Fig.~\ref{fig:retrieval}. The recall rate on consumer-to-shop task is lower than in-shop due to more challenging scenes and body poses.

\noindent
\textbf{Fashion Landmark Detection.}
The performance of landmark detection (\cite{liu2016fashion,yan2017unconstrained}) is represented as the percentage of detected landmarks (PDL), evaluating on different distances from the ground truth, shown as Fig.~\ref{fig:landmark}.

\noindent
\textbf{Fashion Parsing and Segmentation.}
We follow Mask-RCNN~\cite{he2017mask} to construct fashion detection and segmentation framework, using ResNet50-FPN as backbone.We measure the average precision on bounding boxes($AP_{bbox}$) and on masks($AP_{mask}$) in different IoU ratios and object scales. Statistic results are shown in Table~\ref{tab:detect}.

\noindent
\textbf{Fashion Compatibility and Recommendation.}
We use the method in~\cite{vasileva2018learning} to complete fashion compatibility learning that develops different embedding projection strategies. 
Higher Fill-in-the-Blank accuracy and higher compatibility AUC means better performance, shown as Table~\ref{table:learning_comp}.

\section{Impacts and Future Plan}

We released MMFashion v0.1 in November 2019, while continuously maintaining and updating it. MMFashion has already received 324 stars, 21 watches and 66 forks on GitHub in less than half a year. 
In the future, we plan to include not only fashion recognition tasks, but also fashion synthesis tasks, such as virtual try-on. With this endeavor, MMFashion is evolving into a unified platform that eases the users to engage in intelligent fashion systems. We hope this toolkit can help join forces and push forward the development of intelligent models in fashion community.